\documentclass[runningheads]{llncs}
\usepackage{wrapfig}
\usepackage[T1]{fontenc}
\usepackage{graphicx}
\usepackage{bbm}
\usepackage{bm}
\usepackage{amsmath,epsfig}
\usepackage{balance}
\usepackage{booktabs}
\usepackage{graphicx}
\usepackage{cite}
\usepackage{subfigure}
\usepackage{multirow}
\usepackage{hyperref}
\usepackage{float}
\usepackage{caption}
\usepackage{graphicx}
\usepackage{tabularx}
\usepackage{adjustbox}
\usepackage{floatflt}
\usepackage{caption}
\usepackage{tabularx}
\usepackage[misc]{ifsym}
\usepackage{amssymb}
\usepackage{color}

\usepackage{mathptmx}
\begin{document}

\title{MISS: A Generative Pre-training and Fine-tuning Approach for Med-VQA}

\titlerunning{MISS: A Generative Pre-training and Fine-tuning Approach for Med-VQA}
\authorrunning{J. Chen et al.}

\author{Jiawei Chen\inst{1,3}\quad Dingkang Yang\inst{1,3}\quad Yue Jiang\inst{1,3}\quad  \\  Yuxuan Lei\inst{1,3}\quad Lihua Zhang\inst{1,2,3,4}$^{\textrm{\Letter}}$}

%Address and e-mail should NOT be added to the submission paper. They should be present only in the camera ready paper. 
\institute{Academy for Engineering and Technology, Fudan University \and
Engineering Research Center of AI and Robotics, Shanghai, China \and
Cognition and Intelligent Technology Laboratory (CIT Lab) \and
Jilin Provincial Key Laboratory of Intelligence Science and Engineering, Changchun, China
}

\renewcommand{\thefootnote}{}
\footnotetext{$^{\textrm{\Letter}}$Corresponding author.}

\maketitle     

\begin{abstract}
Medical visual question answering\,(VQA) is a challenging multimodal task, where Vision-Language Pre-trained\,(VLP) models can effectively improve the generalization performance. However, most current methods in the medical field treat VQA as an answer classification task which is difficult to transfer to practical application scenarios. Additionally, due to the privacy of medical images and the expensive annotation process, large-scale medical image-text pairs datasets for pretraining are severely lacking. In this paper, we propose an efficient \textbf{M}ult\textbf{I}-task \textbf{S}elf-\textbf{S}upervised-learning-based framework\,(MISS) for medical VQA tasks. Unlike existing methods, we treat medical VQA as a generative task. We unify the text encoder and multimodal encoder and align image-text features through multi-task learning. Furthermore, we propose a Transfer-and-Caption\,(TransCap) method that extends the feature space of single-modal image datasets using Large Language Models\,(LLMs), enabling those traditional medical vision-field task data to be applied to VLP models. We conduct extensive experiments and compare them with existing medical VQA methods adopting a no-generative paradigm. We demonstrate the advantages of pre-training with data generated by the TransCap method and our method achieves excellent results with fewer multimodal datasets. The code has been released at \url{https://github.com/TIMMY-CHAN/MISS.git}.

%ANTS 2024 does not use keywords in the proceedings.
\end{abstract}

\section{Introduction}
\label{sec:intro} 

Visual Question Answering\,(VQA) is a multi-modal task based on vision and language, aiming to provide corresponding answers to the given images and questions. Thanks to the development of Convolutional Neural Networks\,(CNN) and Natural Language Processing (NLP) techniques, some works\cite{10.1007/978-3-030-32251-9_57, 10.1007/978-3-030-87240-3_7, 9802503} have attempted to use CNN and Recurrent Neural Networks (RNN) to extract image and text features respectively for VQA tasks. With the emergence of transformers\cite{vaswani2017attention}, image features and text features can be more easily embedded into the feature space with the same dimension, and VLP models\cite{li2019visualbert, radford2021learning} have emerged continuously and have been proven to be effective solutions for downstream multi-modal tasks. 

While effective, these VLP models have several limitations when applied to medical fields.

Compared with the VQA of natural images, Medical VQA\,(Med-VQA) requires a deeper and more accurate understanding of medical images. At the same time, due to the privacy of medical images and the high cost of high-quality text annotation, large-scale datasets for training Med-VQA are extremely scarce. Therefore, currently, Med-VQA is still a highly challenging task.

Currently, some multimodal models specialized in the medical domain have been proposed, such as M3AE~\cite{chen2022multi}, MRM~\cite{zhang2023large}, and CMITM~\cite{chen2023contrastive}, which unify Masked AutoEncoders (MAE)~\cite{he2022masked} and Masked Language Modeling (MLM) pre-training to learn joint representations of images and texts; MUMC~\cite{li2023masked} utilizes Masked Image Modeling(MIM) by sending masked images to the image encoder as data augmentation; PMC-CLIP builds a new large medical dataset and trains it on a CLIP~\cite{radford2021learning} which pre-trained on natural images. 
However,  the above Med-VQA models still have two key problems:

\textbf{a.} They treat Med-VQA as an answer classification task by selecting the most likely answer from a candidate pool as the output. Such models cannot adapt to diverse questions and be transferred to practical application scenarios, as there are no candidate answers in practical applications. 

\textbf{b.} They utilize image-text pairs crawled from the article centres for pre-training, which contain a lot of noise, and high-quality open-source medical images for other tasks, such as medical image classification and segmentation, have been being ignored.

In this paper, we propose a new pre-training and fine-tuning paradigm for medical image-text tasks, named \textbf{MISS}, and apply it to the Med-VQA task. We treat Med-VQA as an answer-generating task, making our method directly applied to real-world scenarios and generating responses that more closely match human expression. Unlike previous dual-tower multi-modal models, We innovatively unify the text encoder and multi-modal encoder, building a \textbf{J}oint \textbf{T}ext-\textbf{M}ultimodal (\textbf{JTM}) encoder and enabling it to learn joint feature representations using a multi-task learning approach.

To align multi-modal features using unimodal medical images, we propose a novel method called \textbf{Trans}fer and \textbf{Cap}tion (\textbf{TransCap}). This method utilizes unlabeled unimodal datasets to construct image-text pairs, making it the first work in the medical field that combines Large Language Models (LLMs) with unimodal image data to construct multi-modal datasets for visual language pretraining and fine-tuning. We believe that with this pioneering approach, researchers in this field no longer need to be plagued by the lack of relevant high-quality image-text pairs for pretraining. 

Our main contributions are as follows:

$\bullet$ We propose a JTM encoder that escapes extracting text and multimodal features in different stages and enhances the efficiency of joint feature representation extraction.

$\bullet$ We present Transcap, a pioneering method for constructing multimodal medical data based on text-free labeled images and LLMs, which will greatly inspire the construction of pretraining data in medical multimodal fields.

$\bullet$ We introduce a new pre-training and fine-tuning framework named MISS. Not considering the Large-scale Vision-Language models (parameters more than 1B), it is the first pure generative VQA model in the \textbf{medical field}.

\section{Related Work}

\subsection{Medical Visual Question Answering}

The task of Med-VQA is to provide answers based on professional questions posed by the inquirer regarding medical images. In terms of training paradigms, early works\cite{10.1007/978-3-030-32251-9_57, 10.1007/978-3-030-87240-3_7, 9802503, lei2023text} mostly employed RNNs and CNNs to respectively extract textual and visual features. However, these methods often suffer from poor generalization. Thanks to the application of transformers, large-scale pretraining has begun to migrate from the textual domain to the multimodal domain. Training VLP models\cite{li2019visualbert, radford2021learning} using image-text pairs and fine-tuning them for downstream tasks has become the preferred approach for most multimodal tasks. 

In terms of content output, previous works in the Med-VQA field have followed the VQA paradigm in the natural image domain, treating VQA as a classification task\cite{10.1007/978-3-030-32251-9_57, 10.1007/978-3-030-87240-3_7, 9802503, 10.1007/978-3-030-87196-3_20, DBLP:journals/corr/abs-2112-13906}. Specifically, fully connected layers and softmax layers are installed at the output end of the model to calculate cross-entropy loss for all candidate answers. Recently, some works~\cite{li2023masked} have also employed text-based decoders, which calculate MLM loss for all candidate answers and select the answer with the smallest loss as the model's output. This approach is referred to as answer ranking. Although these methods achieve good accuracy on some benchmarks, they still treat VQA as a simple multi-classification task. When transferred to practical tasks without candidate answer pools, these Med-VQA methods cannot be effective. 

In this paper, we propose a pretraining-finetuning paradigm called MISS for Med-VQA tasks. To our knowledge, for \textbf{small-scale} VLMs, this is the first work in the \textbf{medical field} that fully treats VQA as a text-generation task.
\subsection{Visual-Language Pretraining Dataset}

Currently, many work train Med-VQA models with the pretraining-finetuning paradigm. However, the medical field faces a shortage of high-quality image-text pairs for pretraining. ROCO~\cite{pelka2018radiology} collects a large-scale unimodal and multimodal medical dataset from PubMed Central articles and constructs an image-text pairs dataset containing multiple types of medical images by expert radiologists. MediCaT\cite{subramanian-2020-medicat} extracts images and corresponding captions from 131k openly available biomedical papers to construct a dataset containing more than 217k medical images with corresponding captions. However, these methods are similar to those used in the natural image domain to construct multimodal datasets by extracting news images and titles from the internet, and the collected images and captions contain a lot of noise, such as citations, labels, and other irrelevant information. Other high-quality open-source medical images for tasks such as medical image classification and segmentation have been ignored because annotating these images also requires high costs and professional knowledge. 

To deal with the above challenges, we propose an automatic method named TransCap for generating captions for unimodal images. This pioneering method attempts to utilize LLMs to construct a multimodal medical image-text dataset. The image data is clean, and the captions conform to human expression habits.
\section{Method}

\begin{figure*}[htbp]
    \centering
    \includegraphics[width=1\textwidth]{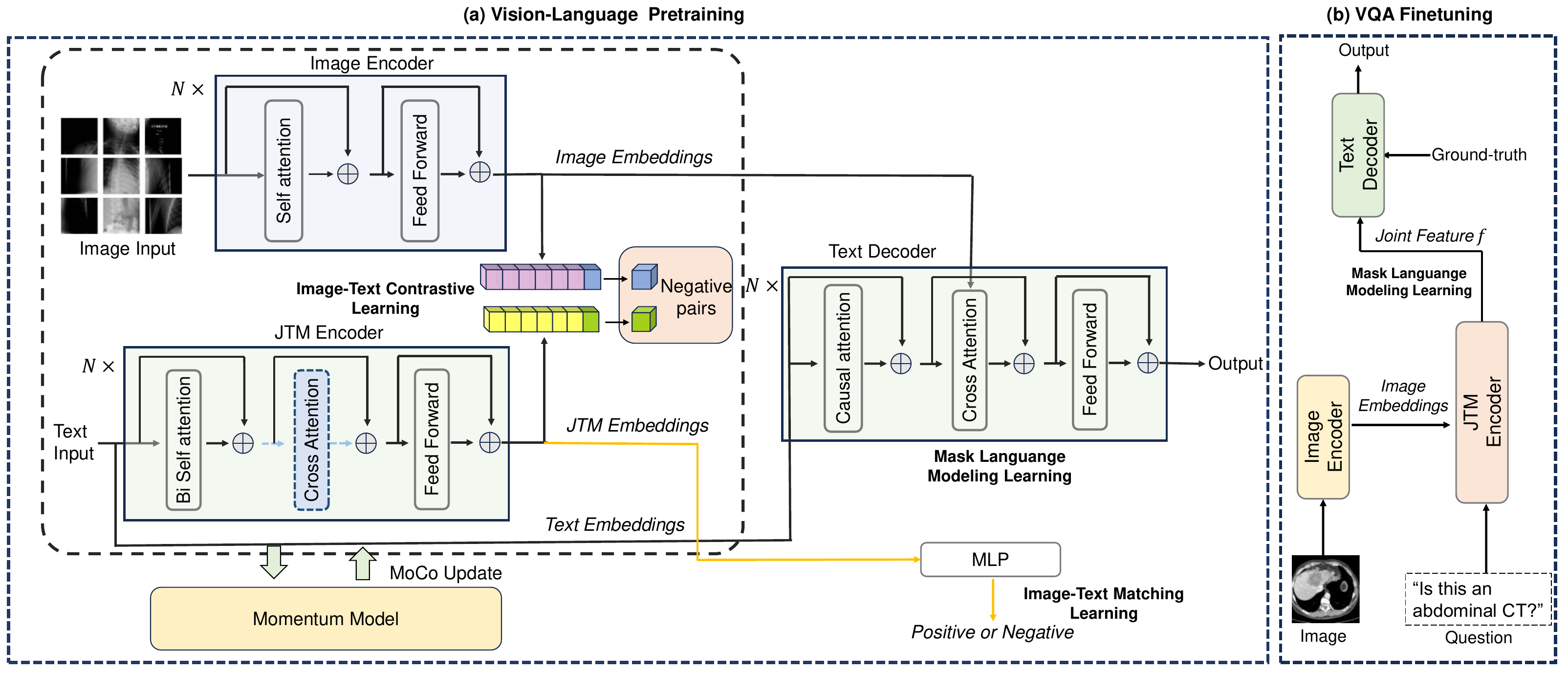}
    \caption{Pretraining (a) and Finetuning (b) of our proposed method. We propose a pretraining and finetuning framework Miss for Med-VQA tasks which is composed of an image encoder, a JTM encoder, and a text decoder. ITC, ITM, and MLM Learning are used for pretraining. In the finetuning stage, the joint feature interacts with tokenized answers for MLM Learning.}
    \label{fig1}
    \vspace{-5pt}
\end{figure*}

\subsection{Overview}

We adopt the pre-training and fine-tuning paradigm for training medical VQA models. In the pretraining stage, we first use image-text pairs to enable the model to learn multi-modal feature representation. In the fine-tuning stage, we use image-question pairs to train the model, enabling it to be applied to Med-VQA tasks ultimately. In the following, we will first introduce our model structure, followed by our pretraining method. Finally, we will present the TransCap method and the implementation details of fine-tuning.
\subsection{Model Architecture}

The architecture of MISS is demonstrated in Figure~\ref{fig1} a, our model adopts the encoder-decoder structure in its entirety. Unlike most dual-tower image-text VLP models in the past, our model's encoder is divided into only two parts - the image encoder and the JTM encoder. For the image encoder, we borrow from the settings adopted by most recent works, which utilize the vision transformer(ViT)\cite{vit} as the image feature extractor. For an image input $I$, it's firstly reshaped into flattened 2D patches and then encoded into a embedding sequence $ \{x_{<cls>},x_{1},...,x_{n}\}$. After that, a 12-layer transformer encoder will extract its high-dimensional features.

The JTM encoder replaces the text and multi-modal encoders used in recent works and it performs representation learning of text and multi-modal features simultaneously. As shown in Figure \ref{fig1}(a), each JTM encoder is composed of 12 transformer-based layers, with each layer containing a bidirectional self-attention layer, a cross-attention layer, and a feed-forward layer. For each text input $T$, it's first pre-processed by the tokenizer into a token sequence. Then, we feed it into the JTM encoder for multi-layer representation learning, where it interacts with the image features through cross-attention. Specifically, we define the text representation as $ \{w_{<cls>},w_{1},...,w_{n}\}$, and the image embeddings are defined as $ \{v_{<cls>},v_{1},...,v_{n}\}$. Both of these representations fuse and compute multimodal representations through 
\begin{equation}
    CrossAttention(Q, K, V ) = SoftMax(\frac{QK^{T}}{\sqrt{d}}+B)V,
\end{equation}
 where text representation $ \{w_{<cls>},w_{1},...,w_{n}\}$ generates query vectors $Q$, and the image representation $ \{v_{<cls>},v_{1},...,v_{n}\}$ generates key vectors $K$ and value vectors $V$.

The decoder part of the model includes a text decoder, which aims to decode the multimodal feature representation obtained by the JTM encoder into an output text representation. The backbone of the text decoder is similar to that of the JTM encoder but replaces the bidirectional self-attention layer in the JTM's per-layer with a causal-attention layer. The text input passes through the causal-attention layer to calculate the text feature representation and then undergoes feature interaction with the multi-modal features through the cross-attention layer. The final features obtained are decoded by the tokenizer to obtain the text output.

\vspace{-6pt}
\subsection{Pre-training}

The pre-training of Vision-Language Models\,(VLMs) aims to align the multimodal features while trying to make the image encoder understand the feature distribution of images in high-dimensional space and comprehend the deep semantic information of medical images. Inspired by METER\cite{meter}, we choose Image-Text Contrastive Learning (ITC), Image-Text Matching (ITM) and Mask Language Modeling (MLM) tasks for multi-modal pretraining. 

To enable the JTM encoder to learn the joint representation of text-multimodal features without being disturbed by the flow of features from another modality during the process of learning one representation, we adopt the method of BLIP\cite{li2022blip} and deform the layer structure of the JTM encoder in different pretraining tasks. Specifically, at the beginning of model pretraining, the distance between $ \{v_{<cls>},v_{1},...,v_{n}\}$ extracted by the image encoder and $ \{w_{<cls>},w_{1},...,w_{n}\}$ of the JTM encoder in high-dimensional space is too far. At this time, it's difficult for the two features to interact and perform ITM and MLM training. Therefore, at the beginning of training, the JTM encoder will discard the cross-attention layer and extract word embeddings so that narrowing the distance between $ \{v_{<cls>},v_{1},...,v_{n}\}$ and $ \{w_{<cls>},w_{1},...,w_{n}\}$ in high-dimensional space through the ITC task. The ITC, ITM, and MLM losses are calculated as delineated below.

\textbf{Image-Text Contrastive Learning} aims to learn unimodal representations before fusion\cite{li2021align}. ITC loss measures the distance of two embeddings in the feature space by a matrix similarity measure $ \mathbf{S} = A^{T}B$. Inspired by MoCo\cite{he2020momentum}, two momentum encoders are created and they respectively have the same architecture as the text encoder and the JTM encoder. Two queues are constructed to store the most recent $M$ image-text representations. The image and text features extracted by the image encoder and JTM encoder are denoted as $e_{V}(v_{cls})$ and $e_{J}(t_{cls})$, and those extracted by momentum encoders are denoted as  $e'_{V}(v'_{cls})$ and $e'_{J}(t'_{cls})$. So we can calculate similarity $\mathbf{S}(I,T) = e_{V}(v_{cls})^{T}e'_{J}(t'_{cls})$ and $\mathbf{S}(T,I) = e_{J}(t_{cls})^{T}e'_{V}(v'_{cls})$, the softmax-normalized similarity between each image-text is calculated as follows: 
\begin{equation}
 \displaystyle{p_{m}^{I2T} = \frac{exp(\mathbf{S}(I,T_{m})/\tau}{\Sigma_{m=1}^{M}exp(\mathbf{S}(I,T_{m})/\tau}, p_{m}^{T2I} = \frac{exp(\mathbf{S}(T,I_{m})/\tau}{\Sigma_{m=1}^{M}exp(\mathbf{S}(T,I_{m})/\tau}}   ,
\end{equation}
where $\tau$ is the temperature parameter.
Similarly, we use the above method to calculate the similarity of the embeddings and their ground truth as $y^{I2T}(I)$ and $y^{T2I}(T)$, the cross-entropy $H$ between $p$ and $y$ is calculated as follows:
\begin{equation}
    \displaystyle{\mathcal{L} = \frac{1}{2}\mathbb{E}_{(I,T)\sim D}[H(y^{I2T}(I),p^{I2T}(I))+H(y^{T2I}(T),p^{T2I}(T)]},
\end{equation}
which is defined as ITM loss $\mathcal{L}_{itc}$.

\textbf{Image-text Matching Learning} is a binary classification task, which measures visual-semantic similarity between images and texts to match and associate them. Following the setting in ALBEF\cite{li2021align}, a linear layer after the JTM encoder is used to predict whether an image-text pair is matched or unmatched given their multimodal feature. The ground-truth label $L$ and the probability of the matched image-text pair $P_{IT}$ are used to calculate the ITM loss:
\begin{equation}
    \displaystyle{\mathcal{L}_{ITM} = \mathbb{E}_{(I,T)\sim D}H(L, P_{IT})}.
\end{equation}

\textbf{Mask Language Modeling Learning} trains the text decoder by randomly masking some tokens in the word vectors. Unlike most VLP models that adopt text decoders that only receive multi-modal features, our decoder simultaneously accepts input from the original tokenized text and the JTM encoder. As shown in Figure \ref{fig1}(a), after the word vectors $ \{w_{<decod>},w_{1},...,w_{n}\}$ undergo causal attention to extract word embeddings, they serve as query vectors $Q$ and interact with image embeddings which generate key $K$ and value vectors $V$ to calculate cross-attention. The MLM loss $\mathcal{L}_{MLM}$ is calculated similarly to that adopted in BERT\cite{devlin2018bert}, where 15\% of the tokens are randomly selected. Then 80\% of selected tokens are replaced with a special token [MASK], 10\% are randomly replaced with other words, and the other 10\% are left unchanged. In our proposed paradigm, while pretraining the model, the tokenized input embeddings $ \{w_{<decod>},w_{1},...,w_{n}\}$ and the image embeddings $\{i_{<cls>},i_{1},...,i_{n}\}$ fuse in the cross-attention layer. Then finetune the joint feature $f$ who fuses with question embeddings $\{q_{<encod>},q_{1},...,q_{n}\}$. Both of them are then sent to the feed-forward layer and calculate the minimized cross-entropy loss between predicted results and ground-truth results:
\begin{equation}
    \displaystyle{\mathcal{L}_{MLM} = \mathbb{E}_{(I,\hat{T})\sim D}H(y^{msk}, p^{msk}(I,\hat{T})},
\end{equation}
where $\hat{T}$ presents a masked token, $p^{msk}(I,\hat{T})$ presents the predicted probability for the masked token, and $y^{msk}$ is a true distribution of vocabulary. 

\subsection{Transfer and Caption}

Transfer and Caption (TransCap) is a method based on Large Language Models (LLMs) to extend the feature space of unimodal image data, which has never been explored in the medical field before. The purpose of TransCap is to overcome the current challenges in medical image-text datasets, which often contain much noise since their images and captions are mostly extracted from open-source papers~\cite{pelka2018radiology}. However, in the medical field, some unimodal tasks, such as medical image classification and lesion segmentation, often have high-quality open-source data. Our method aims to utilize these unimodal datasets to construct high-quality multimodal image-text pairs.

\begin{figure}[t]
    \centering
    \includegraphics[scale=0.29]{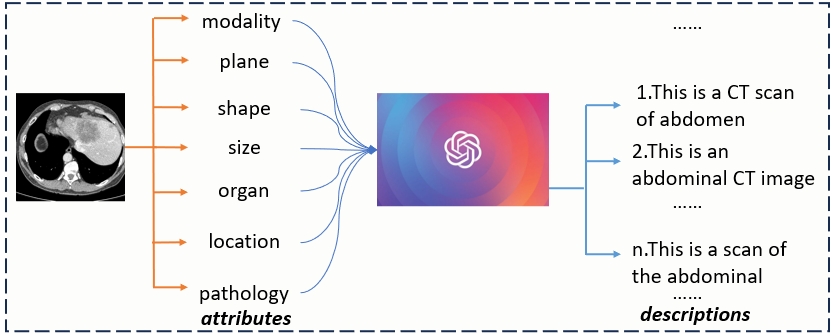}
    \caption{Transfer and Caption unimodal images. We construct image descriptions based on image attributes and ChatGPT.}
    \label{fig2}
    \vspace{-8pt}
\end{figure}

TransCap defines six attributes for medical images: modality, plane, shape, size, organ, location, and pathology. As shown in Figure \ref{fig2}, for each input image $I$, TransCap constructs a corresponding dictionary $dict_{I}\{\}$ with keys representing the six image attributes. For unimodal medical image datasets, TransCap uses ChatGPT to generate attribute content based on dataset information and task labels. Then, it uses this attribute content as a prompt to input the LLM and requests it to generate multiple ways of expressing the attribute's corresponding textual description. The attribute textual description serves as the value corresponding to the attribute key in $dict_{I}\{\}$. During pretraining, TransCap constructs captions by randomly sampling the various attribute contents of the input image dictionary. In this way, we can make use of the previously overlooked large amount of high-quality open-source unimodal image data and obtain captions that are more in line with human expression habits through LLMs. 

For example, the RSNA-PDC~\cite{rsna-pneumonia-detection-challenge} is a chest radiograph (CXR) dataset with a training set of 26,684 CXRs in three categories: Normal, No Lung Opacity/Not Normal, and Lung Opacity. Based on the CXR attributes, the labels of each CXR image can be set to the following types: Modality (indicates data type), Class (can be: Normal, No Lung Opacity/Not Normal, Lung Opacity), Nums (indicates the number of lung opacities), Location (indicates the location of lung opacity).

\begin{figure}[]
    \centering
    \includegraphics[width=1\textwidth]{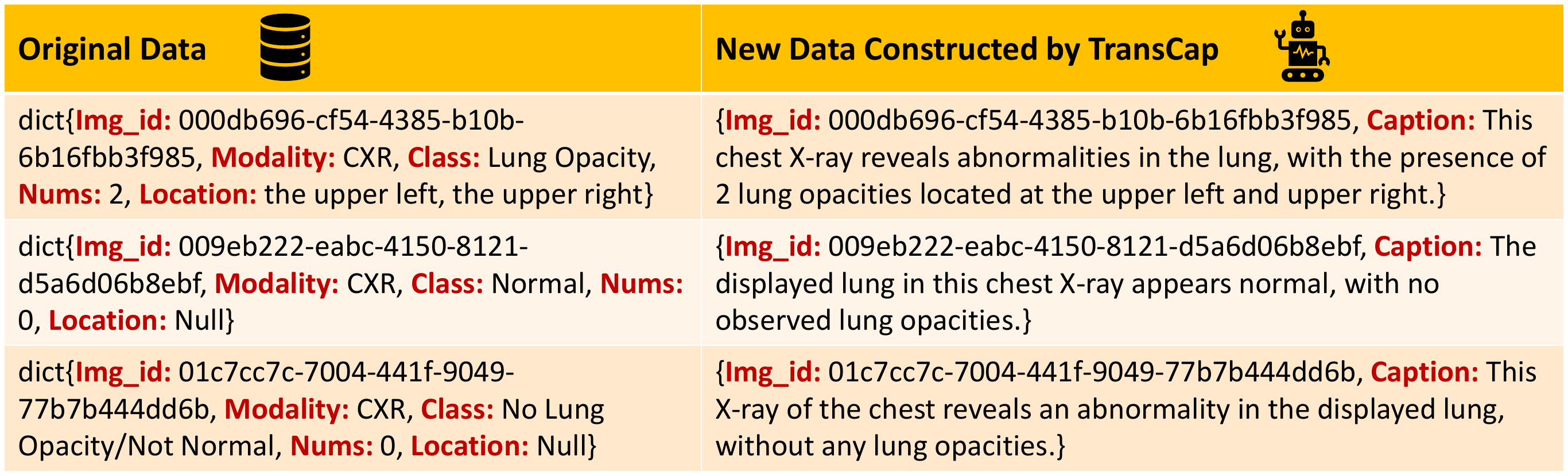}
    \vspace{-12pt}
    \caption{Examples of constructing new image-text pair by TransCap. Discrete image attribute information is converted to image descriptions end-to-end by ChatGPT.}
    \label{fig3}
    \vspace{-8pt}
\end{figure}

The attribute labels of each data will be input into LLM as a prompt, and it is required to generate a caption to describe the attributes of CXR. Figure \ref{fig3} shows a series of examples where the attribute categories of the original single-modal dataset are translated into the caption of an image, the origin data on the left side shows the medical image attributes saved in dictionary form, and the right side shows the image caption constructed by the TransCap method.

\begin{figure}[]
    \centering
    \includegraphics[width=1\textwidth]{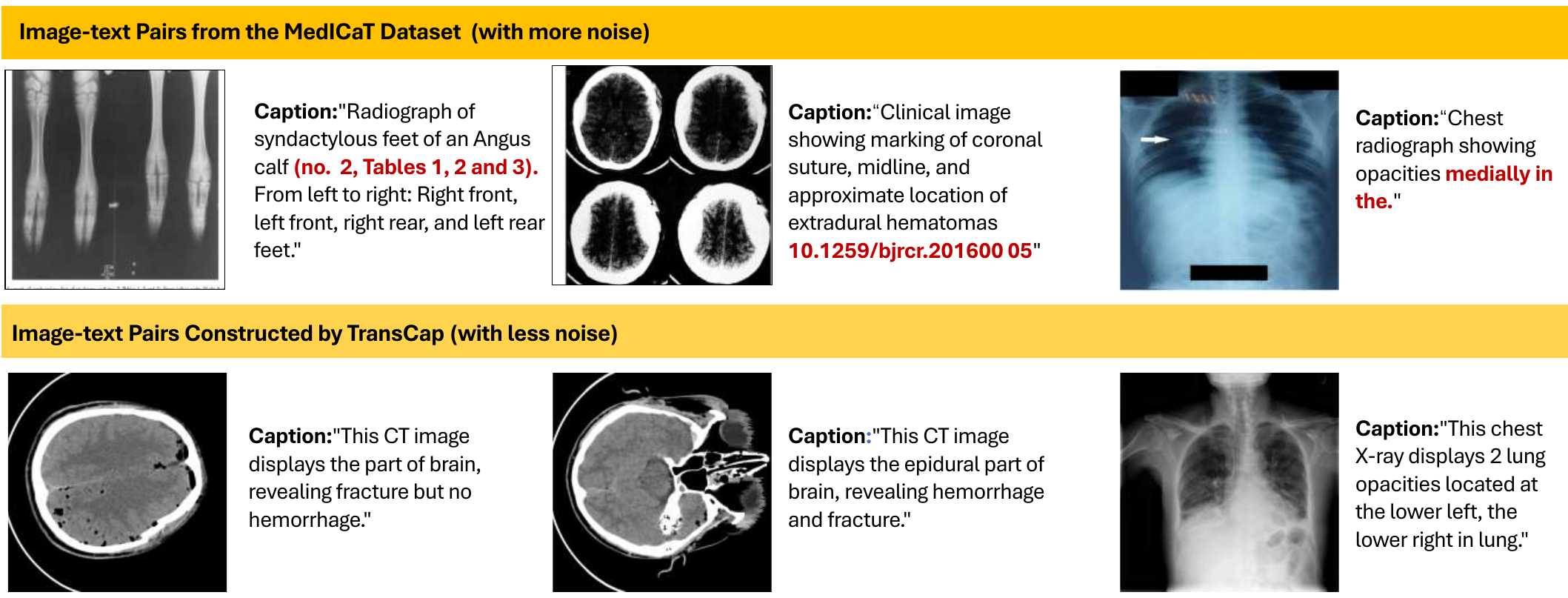}
    \vspace{-8pt}
    \caption{Comparison of data from MedICat Dataset and image-text pair data generated through TransCap. Image-text pairs generated by the TransCap method contain less noise and are more humanized in terms of both image and caption.}
    \label{fig:compare_caption}
    \vspace{-10pt}
\end{figure}

An image from an unimodal dataset is transformed into an image-text pair, and the multimodal data generation is realized with the LLM to obtain a caption set which is more in line with human expression habits. Figure \ref{fig:compare_caption} compares the largest medical multimodal dataset, MedICaT, collected from article centres, with image-text pairs generated by TransCap. One can observe that the image-text pairs gathered from open-source publications contain a lot of noise, such as blurry details and irrelevant captions, whereas those produced by the TransCap method are comparatively less noisy and appear more human-like in both image and caption.

\subsection{VQA Fine-tuning}

As shown in Figure \ref{fig1}(b), during the fine-tuning stage, the image input $I$ undergoes image encoding to extract image embeddings $\{i_{<cls>},i_{1},...,i_{n}\}$. The question input $Q$ is encoded by the JTM encoder to obtain question embeddings $\{q_{<encod>},q_{1},...,q_{n}\}$ and then interacted with the $\{i_{<cls>},i_{1},...,i_{n}\}$ in the cross-attention layer to obtain joint feature representations $f\in \mathbbm{R}^{n+1}$. The tokenized answer is then sent to the causal attention layer to obtain answer embeddings $\{a_{<decod>},a_{1},...,a_{n}\}$, which serve as query vectors in the cross-attention layer. These then interact with the joint feature representations, and the final output is used to calculate LM loss $\mathcal{L}_{LM}$ like Bert with the ground truth.

\vspace{-6pt}
\section{Experiment}

In this section, we compare MISS with a series of previous SOTA medical VQA models. Since most methods treat VQA as a simple classification task or rank task, while our model treats VQA as a generation task, and most methods use different training paradigms and pretraining data scales, a direct result comparison may be unfair for our method. We also conduct extensive ablation studies on our method to demonstrate the effectiveness of the JTM encoder and the TranCap. Next, we will introduce our baseline of the experiments, implementation details, dataset, comparative experiments, and ablation studies respectively.

\subsection{Baseline} 
In this paper, we present an image-text pretraining framework tailored for generative tasks and propose a joint text-multimodal encoder to simultaneously extract features from both images and text through multi-task pretraining. To better compare the advantages of our approach with others, we have constructed a baseline model for our study, which is based on ALBEF\cite{li2021align}. Specifically, in terms of model architecture, the baseline model comprises a ViT-Base as the backbone for the image encoder, consisting of 12 transformer layers; a BERT-based text encoder and multimodal encoder, wherein the first six layers of it serve as the text encoder, which is identical to the original BERT encoder, while the latter six layers incorporate cross-attention between the self-attention and feed-forward layers to function as the multimodal encoder. A BERT-based text decoder is connected to the multimodal encoder and used for causal language modeling.

In terms of the pre-training of the baseline model, it still follows ALBEF and sets up three pre-training tasks: Image-Text Contrastive Learning, Image-text Matching Learning, and Mask Language Modeling Learning. For fine-tuning the VQA task, we still use Mask Language Modeling Learning as the fine-tuning task. Since ALBEF still treats VQA as a RANK task in testing, we modified the output end of the model to enable it to directly generate text.

% \usepackage{multirow}
% \usepackage{multirow}
% \usepackage{multirow}
\begin{table*}[htbp]
\caption{Comparsion with other works which have different methods, training paradigms and types of tasks on ACC (\%). w/o” means the
without.}
\resizebox{\linewidth}{!}{%
\begin{tabular}{lccccccccc}
\hline
\multirow{2}{*}{Methods} &
  \multirow{2}{*}{Training paradigm} &
  \multicolumn{1}{c|}{\multirow{2}{*}{Type of task}} &
  \multicolumn{3}{c|}{VQA-RAD} &
  \multicolumn{3}{c}{SLALKE} \\ %\cline{4-9} 
               &                      & \multicolumn{1}{c|}{}                        & CLOSED        & OPENED & \multicolumn{1}{c|}{OVERALL} & CLOSED & OPENED & OVERALL \\ \hline
\multicolumn{9}{c}{Small-scale Vision-Language Models}                                                                                                                         \\ \hline
MEVF\cite{10.1007/978-3-030-32251-9_57}           & Meta learning        & \multicolumn{1}{c|}{classification}          & 75.1          & 43.9   & \multicolumn{1}{c|}{-}       & -      & -      & -       \\
MMQ\cite{10.1007/978-3-030-87240-3_7}            & Supervised learning  & \multicolumn{1}{c|}{classification}          & 75.8          & 53.7   & \multicolumn{1}{c|}{67}      & -      & -      & -       \\
VQAMIX\cite{9802503}         & Supervised learning  & \multicolumn{1}{c|}{classification}          & 79.6          & 56.6   & \multicolumn{1}{c|}{70.4}    & -      & -      & -       \\
AMAM\cite{PAN2022109763}           & Supervised learning  & \multicolumn{1}{c|}{\textbf{classification}} & \textbf{63.8} & 80.3   & \multicolumn{1}{c|}{73.3}    & -      & -      & -       \\
CPRD\cite{10.1007/978-3-030-87196-3_20}           & Pretraing-finetuning & \multicolumn{1}{c|}{classification}          & 80.4          & 61.1   & \multicolumn{1}{c|}{72.7}    & 83.4   & 81.2   & 82.1    \\
PUBMEDCLIP-MEVF\cite{DBLP:journals/corr/abs-2112-13906} &
  Pretraing-finetuning &
  \multicolumn{1}{c|}{classification} &
  78.1 &
  48.6 &
  \multicolumn{1}{c|}{66.5} &
  76.2 &
  79.9 &
  77.6 \\
M3AE\cite{chen2022multi}           & Pretraing-finetuning & \multicolumn{1}{c|}{classification}          & \textbf{83.4} & 67.2   & \multicolumn{1}{c|}{77}      & 87.8   & 80.3   & 83.2    \\
MTL\cite{cong2022caption}            & Pretraing-finetuning & \multicolumn{1}{c|}{classification}          & 79.8          & 69.8   & \multicolumn{1}{c|}{75.8}    & 86.1   & 80.2   & 82.5    \\
MUMC\cite{li2023masked}           & Pretraing-finetuning & \multicolumn{1}{c|}{ranking}                 & 84.2          & 71.5   & \multicolumn{1}{c|}{79.2}    & -      & -      & 84.9    \\
OURS(w/o Transcap) &
  Pretraing-finetuning &
  \multicolumn{1}{c|}{generating} &
  80.35 &
  \textbf{71.81} &
  \multicolumn{1}{c|}{76.05} &
  82.91 &
  \textbf{81.47} &
  82 \\ \hline
\multicolumn{9}{c}{Large-scale Vision-Language Model}                                                                                                                          \\ \hline
LLaVA(7B)\cite{li2023llava}       & Pretraing-finetuning & \multicolumn{1}{c|}{generating}                 & 65.07         & 50.00  & \multicolumn{1}{c|}{-}       & 63.22  & 78.18  & -       \\
LLaVA-Med(7B)\cite{NEURIPS2023_5abcdf8e} & Pretraing-finetuning & \multicolumn{1}{c|}{generating}                 & 84.19         & 61.52  & \multicolumn{1}{c|}{-}       & 85.34  & 83.08  & -       \\
LLaVA-Med(13B)\cite{NEURIPS2023_5abcdf8e} & Pretraing-finetuning & \multicolumn{1}{c|}{generating}                 & 81.98         & 64.39  & \multicolumn{1}{c|}{-}       & 83.17  & 84.71  & -       \\ \hline
\end{tabular}
}
\label{table1}
\end{table*}

\subsection{Dataset and Metrics}

We consider two Med-VQA benchmarks: the VQA-RAD dataset\cite{lau2018dataset} and the Slake dataset\cite{liu2021slake}. VQA-RAD contains 315 radiology images and 3,515 QA pairs annotated by clinicians, which are evenly distributed over the head, abdomen, and chest. SLAKE is a semantically-labeled knowledge-enhanced dataset for Med-VQA, which consists of 642 radiology images and 14,028 QA pairs created by experienced physicians. VQA-RAD doesn't provide a test set, and we extract 1,797 QA pairs to train the model, and the rest 451 pairs are used to test. For Slake, 14,028 QA pairs are divided into 70\% training, 15\% validation, and 15\% testing subsets.

Considering that existing state-of-the-art\,(SOTA) Med-VQA methods have different task paradigms, we cannot take BELU, BERTScore and et.al metrics to evaluate our model, which is usually adopted to evaluate generative models. To intuitively compare with the existing methods, we still choose accuracy (ACC\,(\%)) as the only evaluation metric, although the evaluation of ACC is not necessarily fair for us compared with the methods using the classification paradigm or ranking paradigm.

\subsection{Comparison with Other Methods}

We conduct a comparative evaluation of our method against the existing SOTA approaches on VQA-RAD and Slake. For small-scale VLMs, our model is the only one that treats Med-VQA as a generative task compared with past research, while others have approached VQA as answer classification or ranking tasks. When we evaluate the ACC, for closed-ended questions with only ``yes'' or ``no'' answers, we utilize automated evaluation methods. For open-ended questions, we follow the approach outlined in \cite{lau2018dataset} and conduct manual evaluations comparing the generated responses with ground-truth answers. We take the model without TransCap as our base model.

Table \ref{table1} demonstrates our comparison with existing methods on the Slake Dataset and the VQA-RAD Dataset. Even if the current Large-scale VLM achieves better results, the extremely low pre-training cost and parameter amount of the small-scale VLM cannot be ignored. Apart from the large-scale VLM LLaVA (7B or 13B), our base model achieves the best accuracy in open-ended questions for Slake,  which reaches 81.47\%, surpassing all methods employing answer classification and ranking tasks. For VQA-RAD, although the test sets selected by each method may vary, the results show that our model still achieves good performance. Table \ref{table3} demonstrates the comparison with those adopting the pre-training and fine-tuning paradigm on the scale of images for pre-training, our model has a smaller pre-trained image scale compared with methods in the pre-trained fine-tuning paradigm, using only 38,800 images. The competitive results demonstrate the efficiency of the JTM encoder.

\begin{figure}[]
    \centering
    \includegraphics[scale=0.4]{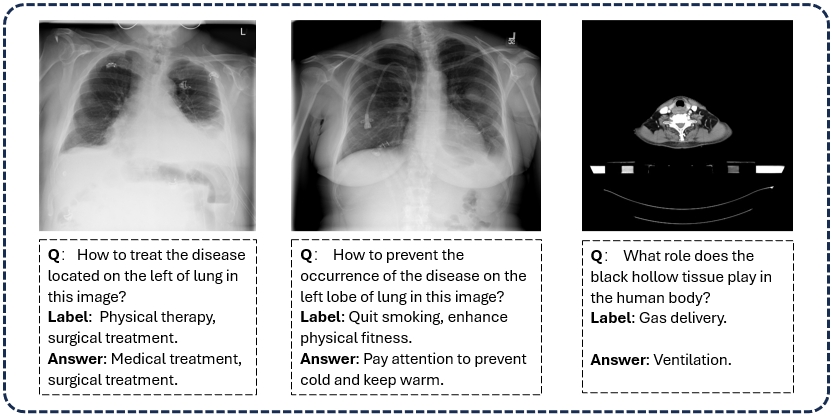}
    \caption{Answers of our method and the ground truth\,(Label).}
    \label{fig5}
    \vspace{-10pt}
\end{figure}

Figure \ref{fig5} showcases a comparison between the responses generated by our generative model and the ground-truth answers for select questions. In some open-ended questions, Our model generates responses that differ from the ground truth but lead to the same destination, this diversity highlighting the advantages of a generative Med-VQA model.

\subsection{Abslation Studies}

\begin{table}[]
\centering
\caption{Comparison of SOTA methods adopting pre-training and fine-tuning paradigm but with different numbers of pre-trained images on open-ended ACC.}
%\resizebox{\linewidth}{!}{%
\begin{tabular}{lc|ccc}
\hline
Methods            & Pre-train \# images & Type of task   & VQA-RAD        & SLALKE         \\ \hline
CPRD               & 22,995            & classification & 61.1           & 81.2           \\
PUBMEDCLIP-MEVF    & 11,779            & classification & 48.6           & 79.9           \\
M3AE               & 298,000           & classification & 67.2           & 80.3           \\
MTL                & 87,952            & classification & 69.8           & 80.2           \\
MUMC               & 387,000           & ranking        & 71.5           & -              \\
OURS(w/o Transcap) & 38,800            & generating     & \textbf{71.81} & \textbf{81.47} \\ \hline
LLaVA(7B)          & -                 & generating     & 50.00          & 78.18          \\
LLaVA-Med(7B)      & 1M                & generating     & 61.52          & 83.08          \\
LLaVA-Med(13B)     & 1M                & generating     & 64.39          & \textbf{84.71} \\ \hline
\end{tabular}
%}
\label{table3}
\end{table}
To demonstrate the effectiveness of different components of our method, we perform an ablation study on Slake, with the results shown in Table \ref{table2}. There are several observations drawn from the results. The model without pre-training achieves only 50.99\% accuracy on closed-ended questions, indicating that it can understand the task type through VQA fine-tuning but cannot fully understand the semantics of medical images. When MISS did not utilize the JTM encoder and conventional multi-modal models were used to set up the text encoder and multi-modal encoder, our global accuracy rate was 1.64\% lower than that of the base model, indicating that the JTM encoder can extract joint features more efficiently.

When our model uses both the JTM encoder and the TransCap method, we compare the impact of TransCap on our model by increasing the amount of pretraining data. As shown in the table, when using the TransCap method, with only an increase of less than 12k pretraining images, our open-ended accuracy and closed-ended accuracy increased by 1.03 and 0.5, respectively, demonstrating the positive effect of TransCap on VQA performance. Since most of the captions generated by TransCap are judgmental statements, the proportion of judgmental captions in the pretraining data continues to increase, resulting in a slight decrease in accuracy on open-ended questions; at the same time, it also leads to a certain increase in overall accuracy, ultimately reaching 83\%.

\begin{table}[]
\setlength{\tabcolsep}{4pt}
\centering
\caption{Ablation studies on different components of our method, ``w/o'' means the without.}
%\resizebox{\linewidth}{!}{%
\begin{tabular}{lc|ccc}
\hline
\multicolumn{1}{c}{\multirow{2}{*}{Methods}} & \multirow{2}{*}{\begin{tabular}[c]{@{}c@{}}Pre-train\\ \# images\end{tabular}} & \multicolumn{3}{c}{SLALKE} \\
\multicolumn{1}{c}{} &        & CLOSED         & OPENED         & OVERALL     \\ \hline
ours (w/o pre-train)   & 0      & 50.99          & 3.82           & 19.6        \\
ours (w/o TranScap)   & 38,800 & 82.91          & 81.47          & 82          \\
ours (w/o JTM)        & 38,800 & 82.82          & 79.11          & 80.36       \\
ours (JTM+TranScap)   & 50,000 & 83.94          & \textbf{81.87} & 82.47       \\
ours (JTM+TranScap)   & 70,000 & 83.94          & 81.44          & 82.38       \\
ours (JTM+TranScap)   & 90,000 & \textbf{84.51} & 81.19          & \textbf{83} \\ \hline
\end{tabular}
%}
\label{table2}
\end{table}

\subsection{Implement Details}
%这里我们将讲述我们预训练和微调模型的实验细节。我们的所有训练在单张NVidia RTX8000-50G显卡上训练在预训练阶段，我们未使用任何数据增强手段；我们使用Adamw优化器，cosine learning rate decay，初始学习率为2e-5，weight decay = 0.05,最小学习率为0，在预训练数据集上共训练100个epoch；此外，预训练阶段，我们模型输入图像的大小为224*224。在微调阶段，我们采用的优化器设置、学习率与预训练阶段相同，输入图像大小为480*480，共训练200个epoch。对于我们的基础模型，vit视觉编码器共包含12层transformer,JTM编码器与文本解码器均包含12层transformer-based layer。
Here, we will present the experimental details of our pre-training and fine-tuning models. All of our training was conducted on a single NVIDIA RTX8000-48GB GPU. During the pre-training stage, we did not use any data augmentation techniques. We used the Adamw optimizer with cosine learning rate decay, an initial learning rate of 2e-5, weight decay of 0.05, a minimum learning rate of 0, and training for 100 epochs on the pre-training dataset. Additionally, during the pretraining stage, the input image size for our model was 224x224 pixels. In the fine-tuning stage, we used the same optimizer settings and learning rate as in the pre-training stage, with an input image size of 480x480 pixels, and trained for 200 epochs. For our baseline model, the Vit-based visual encoder consists of 12 layers of transformer, and both the JTM encoder and text decoder contain 12 layers of transformer-based layers.

\section{Conclusion}
In this paper, we propose a pre-training and fine-tuning framework for Med-VQA tasks. We treat Med-VQA as a generative task and propose a Joint Text-Multimodal encoder and align image-text features through multi-task learning. Furthermore, we propose a Transfer-and-Caption method that extends the feature space of single-modal image datasets using LLMs, enabling the traditional medical vision-field task data to be applied to VLP. We demonstrate excellent results with fewer multimodal datasets and the advantages of generative VQA models through experiments. We hope that our method will encourage the development of Med-VQA in both data and model aspects.

\section{Acknowledgement}
This work is supported in part by the National Key R\&D Program of China (2021ZD011 3503) and in part by the Shanghai Municipal Science and Technology Committee of Shanghai Outstanding Academic Leaders Plan (No. 21XD1430300).

\end{document}